%% file: main.tex
\documentclass[letterpaper, 10pt, conference]{ieeeconf}

\IEEEoverridecommandlockouts
\usepackage{graphicx}
\usepackage{amsmath}
\usepackage{amssymb}
\usepackage{xcolor}
\usepackage{url}
\usepackage{multirow}
\usepackage{indentfirst}
\usepackage{subfigure}
\usepackage[hidelinks]{hyperref}
\usepackage{booktabs}

\title{\LARGE \bf Project SCOUT: Interceptor Drone for Perimeter Defense}

\author{
Azmain Yousuf$^{\dagger*}$, Siwei Cai$^{\dagger}$, Knut Peterson, Lifeng Zhou, and David Han\\
Drexel University\\
{\tt\small \{ay445, sc3568, kp3275, lz457, dkh42\}@drexel.edu}\\
{\small $^{\dagger}$Equal contribution. $^{*}$Corresponding author.}
}

\begin{document}

\maketitle
\thispagestyle{empty}
\pagestyle{empty}

\begin{abstract}
The rapid proliferation of unauthorized unmanned aerial vehicles (UAVs) has created a growing need for robust, jamming-resistant counter-UAV systems for perimeter defense.
This paper presents \textbf{SCOUT} (Spatial Computation for Optimized UAV Tracking), a ROS-integrated onboard perception and control framework for real-time aerial defense against incoming UAVs.
SCOUT performs visual detection, target association, track filtering, and control command generation directly onboard the defender UAV, without relying on external sensing infrastructure or ground-station computation.
To provide stable control inputs, the perception pipeline combines TensorRT-accelerated drone detection with ByteTrack-based association and a lightweight track-retention state machine.
The state machine rejects abrupt target jumps and maintains short-term target continuity during temporary detection degradation, reducing unstable control responses caused by false detections or target switching.
We evaluate the proposed architecture through an integrated hardware deployment executing a planar ``goalkeeping'' interception strategy.
In this setting, the defender UAV tracks the incoming target and adjusts its motion to maintain a blocking configuration near the protected boundary.
Real-world flight results show that SCOUT maintains valid target detections for 92.2\% of frames while operating at real-time onboard detection rates, demonstrating the feasibility of visual tracking and closed-loop control for UAV perimeter defense.
A video demonstration of the end-to-end perimeter defense operation is available online.\footnote{\url{https://youtu.be/OxrbMOGu-Kg?si=WZKJxjPzLXWfQ3sX}} 
\end{abstract}


\input{sections/1.Introduction}

\input{sections/2.Related}

\input{sections/3.Method}
\input{sections/4.Experiment}
\input{sections/5.Conclusion}

\bibliographystyle{ieeetr}
\bibliography{ref}

\end{document}

%% file: sections/1.Introduction.tex
\section{Introduction}

The widespread adoption of highly maneuverable and low-cost unmanned aerial vehicles (UAVs) has introduced new security vulnerabilities to critical infrastructure, civil airspace, and restricted perimeters ~\cite{zhao_2022_antiuav, rouhi2024_LRDD}. While traditional counter-UAV methodologies rely heavily on electronic jamming, spoofing, projectile-based neutralization methods ~\cite{2019_muller_detection}, or missile interceptors, these approaches have proven increasingly ineffective against wire-guided systems and low-cost UAV platforms. As an alternative, defender-UAV-based interception provides a cost effective approach for protecting restricted areas, but it requires reliable onboard perception and control under tight real-time constraints~\cite{2017_goppert_cuas,2022rudys_hostile_uav}.

Compared with passive air-to-ground surveillance, active drone interception requires the perception output to be directly usable by a closed-loop flight controller. The target must be detected, tracked across frames, and used for stable control commands despite rapid relative motion, scale changes, and temporary visual degradation ~\cite{zheng_2021_detfly,2023_kim_chasing}. Even short tracking dropouts or abrupt target-center jumps can produce unstable control inputs, causing the defender to lose track of the target. Achieving real-time closed-loop operation demands tight coupling between visual models and low-level flight control systems, all executing within strict size, weight, and power constraints of an embedded platform \cite{TensorRT}.

In real-world deployment, the performance of theoretically sound computer vision algorithms is often limited by system-integration constraints. Deep neural networks such as the You Only Look Once (YOLO) architecture~\cite{yolo} running on edge accelerators must sustain sufficient inference throughput while avoiding latency spikes during flight~\cite{liu2023edgeyolo}. Standard bounding-box de-duplication steps like Non-Maximum Suppression (NMS) can introduce post-processing bottlenecks or CPU-GPU memory transfer overhead on low-power accelerators~\cite{zhao2024detrs}. Additionally, standard tracking frameworks frequently confuse the target drone with its background elements, resulting in mixed-up tracking labels and broken trajectories \cite{aharon2022bot}, a vulnerability heavily exacerbated by the dense environmental clutter and severe distance scaling typical of long-range drone streams \cite{rouhi2024_LRDDv2}. For a perimeter defense system, target-switching mid-flight would be catastrophic. Therefore, the perception pipeline must provide temporally consistent target-center estimates rather than isolated frame-level detections \cite{zhang2022bytetrack,bewley2016simple}.

To address these deployment challenges, this paper introduces SCOUT (Spatial Computation for Optimized UAV Tracking), an onboard framework for real-time UAV perimeter defense. 
Rather than treating deep-learning-based small object detection ~\cite{khorshand_2025_icce,sahi,zheng2021air,dong2025securing}, and trajectory tracking as independent pre-processing stages, SCOUT couples TensorRT-accelerated detection, ByteTrack-based association, outlier filtering, and autonomous control within a unified onboard pipeline. 
This pipeline outputs temporally consistent target-center errors, which are used by the controller to guide the defender UAV toward blocking against the incoming target.
All perception and control computations are executed onboard the defender UAV, without reliance on ground-station computation. The primary technical contributions of this work are summarized as follows:
\begin{itemize}
    \item We present the end-to-end system integration of the SCOUT framework, demonstrating a robust bridge between high-throughput TensorRT-accelerated \cite{TensorRT} deep inference engines and a real-time ROS control loop executing entirely on an Nvidia Jetson Orin NX.
    
    \item We develop a target-retention and outlier-rejection module that combines ByteTrack-based association with spatial displacement filtering to improve the temporal consistency of target-center estimates used for control.
    
    \item We validate the proposed system through real-world hardware experiments, demonstrating real-time onboard perception, stable target-center estimation, and closed-loop defensive maneuvering for UAV perimeter defense.

\end{itemize}

%% file: sections/2.Related.tex
\begin{figure*}[tbp]
    \centerline{\includegraphics[width=1\linewidth]{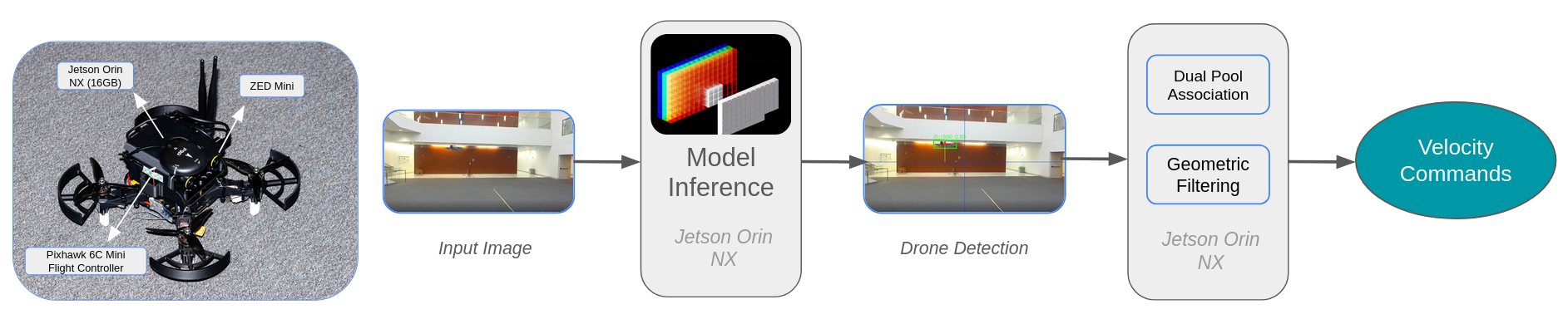}}
    \caption{SCOUT Visual Perception Pipeline: Images from the defender UAV are processed by a TensorRT-accelerated detection model on the Jetson Orin NX. The resulting drone detections are temporally associated and geometrically filtered before being converted into velocity commands for closed-loop interception control.}
    \label{fig:inference_pipeline}
\end{figure*}

\section{Related Work}

\subsection{Drone Detection Datasets}

Proliferation of drone technologies for personal, commercial and governmental uses have led to the creation of several datasets for drone detection studies. One of the early works in this area, the DroneHunter \cite{wyder_2019_dronehunter} dataset, focused on detection of drones for GPS-denied scenarios. The dataset contains 58k images, 10k of which were generated in simulation, and the others were collected in both indoor and outdoor environments.
Other datasets, such as Det-Fly \cite{zheng_2021_detfly} and DUT Anti-UAV \cite{zhao_2022_antiuav} focus mainly on outdoor drone detection in varying conditions. Det-Fly was one of the first datasets to consist of mainly 4K resolution images, and includes a total of 13k images for drone detection. DUT Anti-UAV contains 10k images, with the main highlight being the 35 different drone models used in the dataset. It additionally contains a variety of weather and lighting conditions.

The LRDD series of datasets~\cite{rouhi2024_LRDD,rouhi2024_LRDDv2,peterson2026_LRDDv3}  focused mainly on long-range drone detection, and in addition to the diverse weather and lighting conditions captured in the dataset, is especially useful due to its scale. In total, the dataset series is composed of over 140K annotated images. Other drone detection datasets of similar scale were mainly from static cameras, so LRDD is unique in providing large-scale data from a mobile view. Crucially, while these existing datasets heavily support passive computer vision and detection benchmarking, they do not provide data collected from or structured for closed-loop, onboard interception or dynamic control evaluation.

\subsection{Drone Detection and Tracking Models}
In addition to datasets, several models have been very effective at drone detection and tracking. One such model is YOLO \cite{yolo}, which provides a flexible framework for training models and options for different model sizes based on hardware requirements. YOLO also has many versions, with continuous improvements to the current model \cite{Jocher_Ultralytics_YOLO_2026}. Recently, transformer-based methods such as DETR \cite{carion2020detr} have also been shown to be effective for detection tasks, with potential applications to drone detection. Building on baseline detection models, tracking algorithms provide additional capabilities for following object movements in video. Some detection models allow for easy integration with recent trackers, and YOLO natively supports the BoT-SORT \cite{aharon2022bot} and ByteTrack \cite{zhang2022bytetrack} tracking algorithms.

While many of these detection approaches are generic, some work has been done to specifically explore methods for drone detection. Khorshand et. al. \cite{khorshand_2025_icce} tested the SAHI \cite{sahi} approach to small object detection for detecting drones, showing that it was effective at boosting long-range detection performance. Other detection strategies for drones have also been effective, such as a coarse-to-fine detection strategy based on vision transformers \cite{rebbapragada20204_course_to_fine}, or background estimation and structural adaptive change detection for counter-UAV applications \cite{2019_muller_detection}.
While this body of work achieves high accuracy on static test splits\cite{2023_kim_chasing}, strong standalone detection and tracking benchmarks do not necessarily translate to stable real-time flight control, as raw output fluctuations and bounding box fragmentation degrade control loop inputs.

\vspace{-1mm}
\subsection{End-to-End Drone Interception Systems}

To translate isolated target localization into active defense capabilities, recent literature has investigated the integration of detection, tracking, and localized guidance into complete end-to-end Counter-Unmanned Aerial Systems(C-UAS).

Early architectures leaned heavily on multi-sensor payloads and ground-centralized processing. Goppert et al. \cite{2017_goppert_cuas} established an early air-to-air C-UAS framework that utilized external radar tracking telemetry to guide a defender drone toward an intruder, deploying electronic jamming mechanisms for neutralization. However, this implementation relied on offboard computation and external radar infrastructure, introducing acute transport delays and rendering it ineffective in infrastructure-less or communication-denied contexts. Similarly, Rudys et al. \cite{2022rudys_hostile_uav} evaluated a multi-UAV neutralization system combining visual tracking with radio frequency (RF) sensors; yet, the tracking loops were fundamentally constrained by ground-station telemetry dependency, limiting the system's dynamic agility when resolving rapid target maneuvers in the near field. 

A notable exception in physical validation is the work by Wyder et al. \cite{wyder_2019_dronehunter}, who successfully deployed an autonomous drone hunter operating via deep learning and entirely onboard computation within GPS-denied environments. Their platform utilized a monocular camera, a localized Tiny YOLO object detector, and a visual-servoing control framework to execute physical air-to-air tracking maneuvers at target velocities up to $1.5\text{~m/s}$. Despite this physical validation, the system revealed critical real-world bottlenecks: inference throughput was severely capped at $8\text{~FPS}$, and visual clutter degraded detection accuracy to $77\%$, inducing frequent track loss and trajectory instability. 

In summary, existing literature provides standalone datasets, high-accuracy detectors, generic tracking paradigms, and isolated interceptor demonstrations. However, many existing frameworks suffer from a strict dependency on external sensors or rigid radar and RF infrastructure \cite{2017_goppert_cuas, 2022rudys_hostile_uav}, a reliance on offboard ground-station computation \cite{2023_kim_chasing}, or severely limited onboard inference speed and tracking robustness when deployed entirely at the edge \cite{wyder_2019_dronehunter}. Consequently, there remains a critical research gap: a unified, modern embedded onboard pipeline that connects high-rate edge inference, robust tracking data association, and deterministic geometric outlier filtering directly to autonomous flight control. The SCOUT platform fills this gap by introducing a fully onboard, edge-accelerated perception and state-machine architecture validated in real-world hardware experiments for active goalkeeper perimeter defense.

%% file: sections/3.Method.tex
\section{Method}

SCOUT is designed to convert onboard visual observations of an incoming UAV into stable control commands for real-time perimeter defense. The system follows a sequential perception-to-control approach, where a drone detector is first trained and fine-tuned for the target UAV, then deployed through a low-latency onboard inference pipeline, as shown in Fig.~\ref{fig:inference_pipeline}. The resulting detections are temporally associated and filtered to produce stable target-center errors, which are finally used by the interceptor controller to generate flight commands. 

\subsection{Detector Training and Fine-tuning}
The first component of SCOUT is a visual detector for finding the target UAV. To achieve accurate drone detection we trained a YOLO26n \cite{Jocher_Ultralytics_YOLO_2026} model using four existing drone detection datasets and fine-tuned on additional data we collected of our target drone. The four datasets we used for pretraining are DroneHunter \cite{wyder_2019_dronehunter} (58.6k images), Det-Fly \cite{zheng_2021_detfly} (13.3k images), DUT Anti-UAV \cite{zhao_2022_antiuav} (10k images), and LRDDv3 \cite{peterson2026_LRDDv3} (102.5k images), which cover a wide variety of drone types, backgrounds, and lighting conditions. They also include indoor and outdoor data and combine to a total of 184.4k images. With these four datasets we trained the YOLO26n model for 50 epochs using the AdamW optimizer with a learning rate of 1e-3 and a batch size of 48 on an AMD Ryzen Threadripper 3960X 24-Core Processor and an NVIDIA GeForce RTX 3090 GPU.
For fine-tuning, we collected and labeled an additional 3,931 images of the DJI Neo target drone, using 3,494 images as a training set and 437 images as a validation set. We then fine-tuned the pretrained model for 10 epochs on our collected data, using the AdamW optimizer with a learning rate of 1e-3 and a batch size of 48 on the same hardware used for pretraining. 

The fine-tuned detector provides the frame-level target localization used by the rest of the SCOUT pipeline. However, frame-level detections alone are not sufficient for closed-loop flight control, since the controller requires fresh and temporally consistent error estimates. We therefore deploy the detector inside a real-time onboard perception pipeline.

\subsection{Real-Time Perception and Inference Pipeline}

The SCOUT visual perception pipeline provides low-latency target-center error estimates under strict embedded computational and latency constraints to drive real-time control commands \cite{2017_goppert_cuas}, \cite{wyder_2019_dronehunter}. The architecture is structured into four decoupled processing layers to reduce input ingestion latency and short-term tracking discontinuities.

\subsubsection{\textbf{Asynchronous Frame Ingestion Management}}
Because the controller should react to the most recent target position rather than outdated frames, the image-ingestion layer prioritizes frame freshness over processing every received image. To shield the low-level flight controller from transport delay variations and queue-induced latency spikes, the frame ingestion layer employs an asynchronous drop-frame strategy. Image data received via ROS topics is decoded and written directly to a strict single-element, thread-safe queue. If an inference cycle is actively executing when a new frame arrives, the existing queue item is immediately overwritten. This ensures that the downstream detection thread always operates exclusively on the most recent spatial frame.

\subsubsection{\textbf{TensorRT Inference Acceleration}}


While the YOLO26n detection model is relatively lightweight, when deploying on an edge device that is also running a full drone flight control stack, it is necessary to further optimize for speed. To accomplish this, we use TensorRT FP16 quantization to further speed up the model and lower its load on the system. When deploying with TensorRT, the model graph undergoes an ahead-of-time compilation pipeline. The PyTorch network architecture is first exported to an intermediate Open Neural Network Exchange (ONNX) representation. The TensorRT builder subsequently parses this graph to generate a hardware-matched binary engine optimized specifically for the embedded Jetson system-on-chip (SoC) architecture \cite{TensorRTInference}, \cite{TensorRTYolov9}. During engine generation, the builder applies structural graph optimizations, including vertical and horizontal layer fusion. By collapsing consecutive convolutions, bias additions, and activation functions into single, unified execution kernels, the engine minimizes memory bandwidth bottlenecks and drastically reduces kernel launch overheads \cite{TensorRT}, \cite{DroneforOil}. 

Acceleration is further augmented by precision reduction to a quantized FP16 execution format \cite{Quantization}. This precision constraint allows the underlying execution engine to route matrix multiplication routines directly to the hardware-level parallel execution architectures of the platform's Tensor Cores rather than standard CUDA vector units \cite{TensorCore}. The dense parallelization maximizes hardware-level throughput while preserving the visual sensitivity required for reliable target localized tracking.

All runtime measurements are collected on the NVIDIA Jetson Orin NX 16GB in maximum-performance mode using a 720p image stream, detector input size of 736 pixels, and batch size of 1. The reported timing covers the full detector pipeline, including ROS frame retrieval, pre-processing, inference, post-processing, target association, geometric filtering, and detection-message publication.
Compared with the initial implementation of PyTorch, the compiled TensorRT engine increases the full performance of the pipeline by approximately 114\%, increasing the average throughput from 22.5~fps to 48~fps. This lower processing time allows for faster flight controller response time, and better tracking performance.


\subsubsection{\textbf{ByteTrack Dual-Pool Data Association}}

Although the detector provides target boxes at each frame, isolated detections can still flicker or disappear under motion blur, scale changes, or partial occlusion. To improve temporal consistency, SCOUT integrates ByteTrack to help maintain consistent tracking across frames \cite{zhang2022bytetrack, aharon2022bot}. While traditional multi-object trackers enforce a rigid, high confidence threshold that discards degraded or blurred detections \cite{zheng_2021_detfly, zhao_2022_antiuav}, ByteTrack uses a dual-pool association strategy, with high-confidence and low-confidence pools, split at a confidence threshold $0.50$. ByteTrack then partitions detections into a high-confidence pool ($S_t \ge 0.50$) and a low-confidence pool ($S_{min} \le S_t < 0.50$), and then matched with previous tracks via a localized Kalman filter for high-confidence detections, or a secondary association step for low-confidence detections. For SCOUT we set a minimum detection confidence threshold of $S_{min} = 0.06$. This low threshold is intentionally selected to retain weak candidate detections that may still correspond to the target during short-term visual degradation.




\subsubsection{\textbf{Outlier Rejection by Sequential Geometric Filtering}}
The final perception stage filters the associated detections to suppress abrupt target jumps before they are converted into control errors. The final layer of the pipeline governs tracking states through a custom sequential filtering state machine designed to isolate transient anomalies and environmental noise. Validated tracks pass through a dual-stage verification framework. First, a geometric jump discontinuity filter evaluates candidate boxes against historical track states.

All bounding-box centers, widths, and heights are normalized by the image width and height. Let $C_t = (c_x^{(t)}, c_y^{(t)})$ be the spatial center of the candidate bounding box at frame $t$, and $C_{t-1} = (c_x^{(t-1)}, c_y^{(t-1)})$ be the center of the last validated track state. The Euclidean displacement $d_t$ is formulated as:
\begin{equation}
d_t = \sqrt{(c_x^{(t)} - c_x^{(t-1)})^2 + (c_y^{(t)} - c_y^{(t-1)})^2}
\end{equation}

The bounding box diagonal $D_{t-1}$ of the preceding verified frame with width $w_{t-1}$ and height $h_{t-1}$ is expressed as:
\begin{equation}
D_{t-1} = \sqrt{w_{t-1}^2 + h_{t-1}^2}
\end{equation}

A candidate box is flagged as a spatial outlier and rejected if it satisfies the joint condition:
\begin{equation}
(d_t > 2D_{t-1}) \land \left(S_t \le \max_{i=1,\ldots,K} S_{t-i}\right)
\end{equation}
where $S_t$ is the current detection confidence score, and the right-hand term represents the maximum historical confidence within a rolling $K=10$ frame buffer. This heuristic rejects low-confidence detections that are spatially inconsistent with the previously accepted target state.

Second, a temporal tracking-loss counter handles complete visual dropouts. If no valid detection is received for $T_{\mathrm{stale}}=0.5$~s, the retained target state is cleared and the tracker is allowed to reinitialize. The output of this filtering stage is a target-center estimate that is sufficiently stable to be used as the input to the interceptor controller.

\subsection{Interceptor Control}
\label{sec:interceptor_control}

The controller maps the filtered image-plane target center into velocity commands for the defender UAV. The interceptor is controlled as a planar goalkeeper rather than a full 3D pursuer. It is constrained to move only in the defender's lateral-vertical motion plane, while the forward velocity and yaw-rate commands are set to zero. This keeps the defender near the protected boundary while allowing it to move toward a blocking configuration against the incoming target.

\begin{figure}[htbp]
    \centerline{\includegraphics[width=0.9\linewidth]{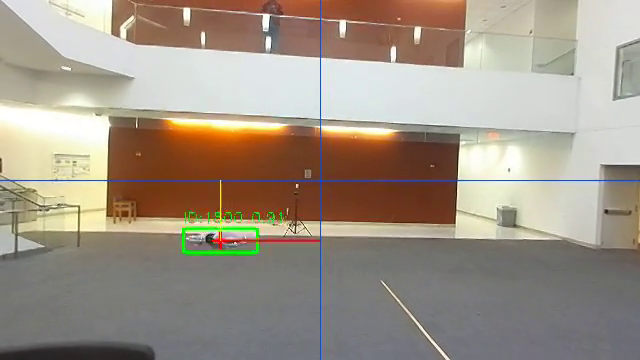}}
    \caption{First-person view from the defender UAV. The detected target UAV is outlined in green, and the image-plane target-center offsets used for control are shown by the red and yellow markers.}
    
    \label{fig:defender_view}
\end{figure}

Let $(c_x,c_y)$ be the normalized bounding-box center, with the image center at $(0.5,0.5)$. 
We define the image-plane servoing errors as
\begin{equation}
    e_u = c_x - \tfrac{1}{2}, \qquad
    e_v = c_y - \tfrac{1}{2}.
    \label{eq:image_error}
\end{equation}
The interceptor then applies a saturated PD law in the local lateral-vertical plane:
\begin{equation}
\begin{aligned}
    v_x &= 0, \qquad \omega_z = 0, \\
    v_y &= -\mathrm{clip}\!\left(k_{p,y}e_u + k_{d,y}\dot{e}_u,\; v_y^{\max}\right), \\
    v_z &= -\mathrm{clip}\!\left(k_{p,z}e_v + k_{d,z}\dot{e}_v,\; v_z^{\max}\right),
\end{aligned}
\label{eq:interceptor_control}
\end{equation}
where $\mathrm{clip}(a,\bar{a})=\min(\max(a,-\bar{a}),\bar{a})$ limits each velocity command to the corresponding flight envelope. The lateral command $v_y$ is expressed in the defender body frame, and $v_z$ is applied along the vertical axis. Fig. \ref{fig:defender_view} shows an example of the defender view with $e_u$ and $e_v$ shown by the red and yellow markers.
With the chosen sign convention, a rightward image error produces a negative body-lateral command, corresponding to rightward motion of the defender, while a downward image error produces a negative vertical command, corresponding to descent.
Before applying the PD law, small image errors are suppressed by dead-bands to prevent detection noise near the image center from causing oscillatory commands.

Control commands are generated only when the vehicle is in OFFBOARD mode and the controller is enabled.
In OFFBOARD mode, the flight controller accepts external velocity commands from the onboard companion computer through MAVROS.
If the latest detection becomes stale, the commanded velocity is set to zero. The vertical command is additionally clamped when the vehicle reaches the prescribed altitude bounds before the command is published to the flight controller through MAVROS. 

%% file: sections/4.Experiment.tex
\section{Experimental Results}

Real-world hardware experiments were designed to verify whether the SCOUT platform can provide real-time target-center estimates and support closed-loop perimeter-defense maneuvers without ground-station computation.
We first describe the experimental UAV platforms and test environment, then report detection-stream performance from a representative flight sequence and discuss the observed system behavior.

\subsection{Experimental Platform} 
\label{sec:experimental_platform} 

\begin{figure}[htbp]
    \centerline{\includegraphics[width=0.9\linewidth]{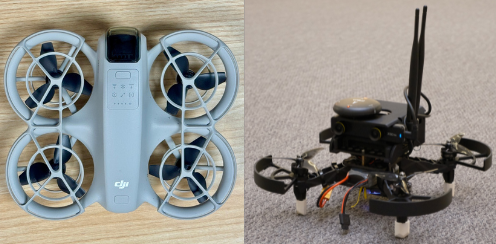}}
    \caption{Invader Drone: DJI Neo (left) and Defender Drone: Custom Built (right)} 
    \label{fig:Defender}
\end{figure}

For this experiment, we use the DJI Neo as the target/adversarial drone. Its compact form factor and fully enclosed propeller guards make it suitable for controlled close-proximity and potential collision testing. The defender drone is a custom-built quadrotor platform designed for onboard perception, planning, and inference. As shown in Fig.~\ref{fig:Defender}, the vehicle is equipped with a Holybro Pixhawk 6C Mini flight controller for low-level flight stabilization and MAVLink-based control. An NVIDIA Jetson Orin NX 16GB module is used as the onboard companion computer, providing sufficient computational capability for real-time perception, planning, and neural-network inference. For visual sensing, the platform uses a ZED Mini stereo camera, which provides visual-inertial odometry for state estimation during flight. 

The software stack runs on ROS 1. MAVROS is used as the interface between the onboard computer and the Pixhawk flight controller, enabling the planner to send motion commands and receive vehicle state feedback through MAVLink. The ZED Mini camera is used to provide VIO-based pose estimates, which are integrated into the ROS-based planning pipeline. High-level planning and decision-making are executed in ROS, while the learned inference module runs fully onboard on the Jetson Orin NX. This configuration allows the defender drone to perform perception, state estimation, planning, and inference without relying on an external ground-station computer during deployment. Overall, this hardware and software configuration enables fully onboard autonomous operation of the defender drone during real-world interaction experiments.

\subsection{Experimental Setup and Protocol}

\begin{figure}[htbp]
    \centerline{\includegraphics[width=0.9\linewidth]{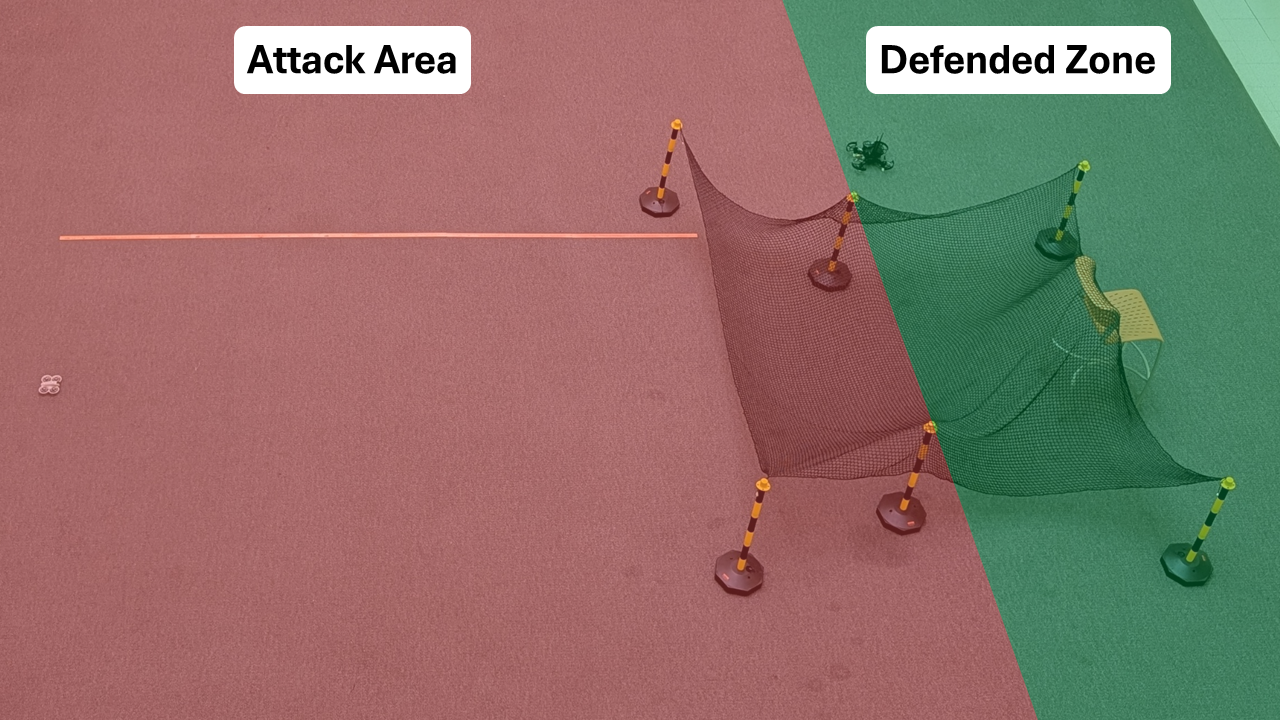}}
    \caption{Overhead view of the indoor perimeter-defense test setup. The defender UAV operates near the protected boundary, while the manually piloted target UAV approaches from the attack side.}
    \label{fig:test_setup}
\end{figure}

Experiments are conducted in an indoor flight volume of approximately $20 \times 40 \times 5$~m as shown in Fig.~\ref{fig:test_setup}. The defender UAV is initialized near the center of the protected boundary at an altitude of approximately $1.2$~m. During each trial, the defender operates autonomously using the SCOUT perception and control pipeline, while the target UAV is manually piloted from the attack side toward the protected region. The target approaches approximately head-on, with additional lateral and vertical motion introduced by the pilot to test whether the defender can maintain visual tracking under changing relative position.

The defender follows a goalkeeper-style behavior. It does not pursue the target through full 3D space, but instead adjusts its position to remain between the incoming target and the protected region. A trial is considered qualitatively successful when the defender maintains a blocking posture and prevents the target from entering the defended region without any physical collision. Because this outcome depends on manual target piloting and close-proximity safety intervention, we do not use binary trial success as the primary quantitative metric. Instead, we conduct multiple hardware trials to validate repeatability of the closed-loop behavior and report detailed perception-stream statistics from a representative 104.53~s flight sequence.

For safety, the defender's commanded velocity was limited to $v_y^{\max}=1.6$~m/s laterally and $v_z^{\max}=2.0$~m/s vertically during all trials.
Under these velocity limits, the defender maintained visual tracking during manually piloted head-on approaches with lateral and vertical target motion.

\subsection{Test Results}
Before analyzing in-flight behavior, we evaluate our fine-tuned detector offline. Table~\ref{table:yolo_performance} contrasts its performance against the 4-dataset pre-trained baseline using our DJI Neo validation set. While the baseline generalizes reasonably well, target-specific fine-tuning substantially improves performance, increasing mAP@50 from 0.878 to 0.955. This performance leap is primarily due to domain adaptation, as the pretraining data did not include the DJI Neo drone model, and our indoor test environment introduces distinct lighting conditions.

\begin{table}[htbp]
\caption{YOLO26n model performance for DJI Neo drone detection.}
\centering
\label{table:yolo_performance}
\begin{tabular}{lcc}
\toprule
\textbf{Model Variant} & \textbf{mAP@50} & \textbf{mAP@50-95} \\ 
\midrule
YOLO26n (pretrained) & 0.878 & 0.600 \\ 
YOLO26n (fine-tuned) & \textbf{0.955} & \textbf{0.824} \\ 
\bottomrule
\end{tabular}
\end{table}

To better understand the impact of each component of the perception pipeline  on tracking accuracy, the system was benchmarked across a continuous sequence of 1,827 frames. As shown in Table~\ref{tab:ablation_filtering}, lowering the baseline detector threshold to $0.06$ allows the standalone YOLO model and the ByteTrack association module to capture a high number of detections, but at the cost of introducing 30 false positives. By introducing the sequential geometric filter, the pipeline completely eliminates these false positives (dropping from 30 to 0) by rejecting spatially inconsistent anomalies. This zero-false-positive performance is achieved with a negligible impact on true detections, sacrificing only a single true positive track during transient boundary conditions and ensuring a highly stable input for the downstream flight controller.

Following this component validation, our quantitative system-level evaluation focuses on whether the onboard perception stream provides sufficiently frequent and stable target-center estimates for closed-loop control. Table~\ref{tab:real_detection_stream} summarizes the detector output topic during the representative real-world flight sequence.
The quantitative evaluation focuses on whether the onboard perception stream provides sufficiently frequent and stable target-center estimates for closed-loop control. Table~\ref{tab:real_detection_stream} summarizes the detector output topic during the representative real-world flight sequence.

\begin{figure*}[t]
    \centerline{\includegraphics[width=0.75\linewidth]{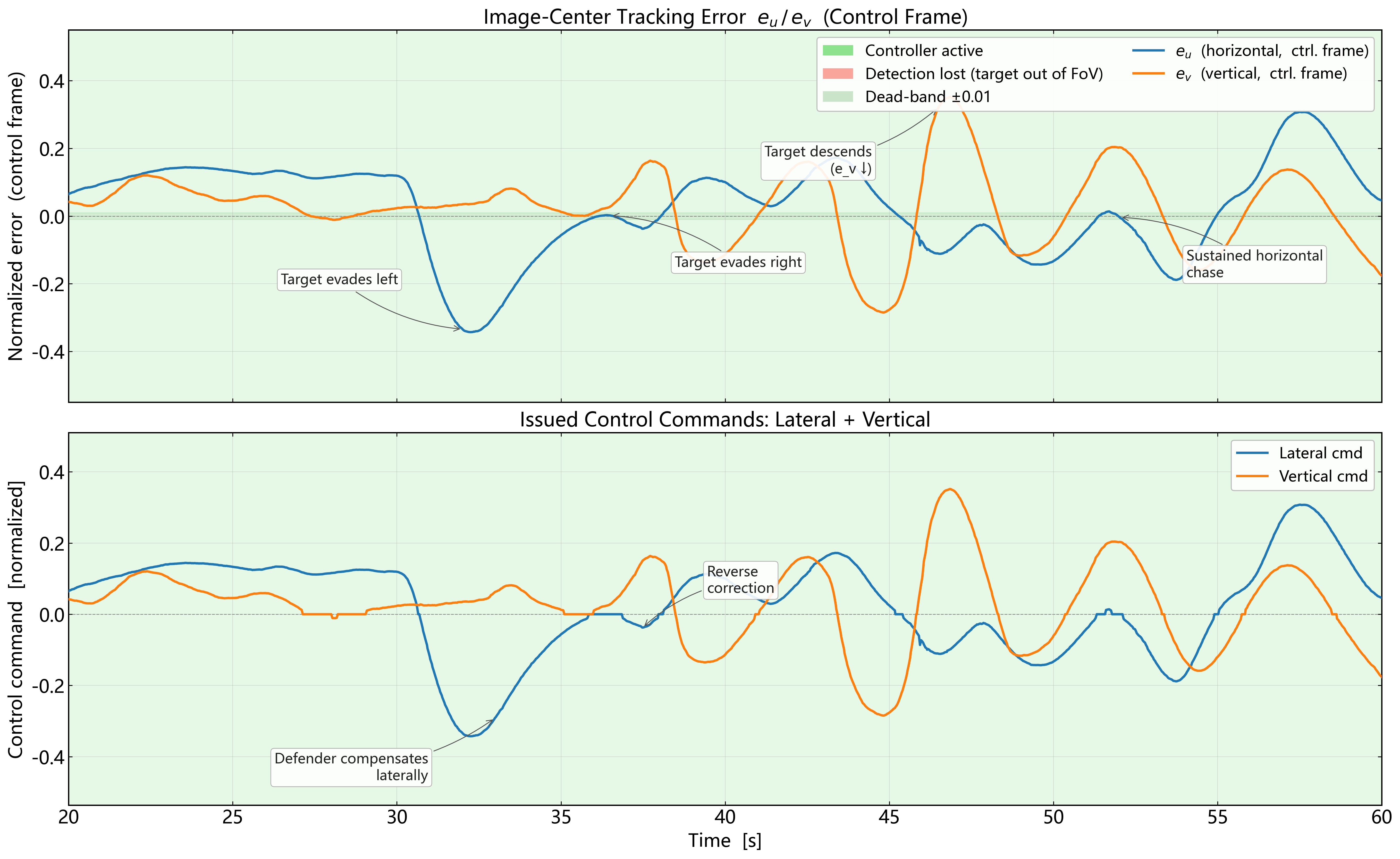}}
    \caption{Closed-loop image-center tracking performance during the active defense phase. The top plot shows the horizontal and vertical tracking errors in the control frame, while the bottom plot shows the corresponding issued lateral and vertical control commands. The defender maintains visual lock with RMSE values of $0.134$ horizontally and $0.107$ vertically in normalized image coordinates.}
    \label{fig:closed_loop_error}
\end{figure*}

To further evaluate closed-loop behavior, we analyze a representative active-defense interval from $t=20$~s to $t=60$~s, where the target performs clear lateral and vertical maneuvers while remaining within the defender's field of view.
As shown in Fig.~\ref{fig:closed_loop_error}, the defender maintains visual lock during this interaction, with RMSE values of $0.134$ horizontally and $0.107$ vertically in normalized image coordinates over the selected interval.
The issued lateral and vertical commands follow the corresponding tracking-error trends, indicating that the perception output is actively coupled to the controller during closed-loop defensive motion.

\begin{table}[htbp]
\centering
\caption{Ablation Study of Tracking and Filtering Components}
\label{tab:ablation_filtering}
\resizebox{\columnwidth}{!}{%
\begin{tabular}{lccc}
\toprule
\textbf{Configuration} & \textbf{Total Frames} & \textbf{Detections} & \textbf{False Positives} \\ \midrule
YOLO (Baseline) & 1,827 & 1,195 & 30 \\
YOLO + ByteTrack & 1,827 & 1,195 & 30 \\
\textbf{YOLO + ByteTrack + Filter} & \textbf{1,827} & \textbf{1,164} & \textbf{0} \\ \bottomrule
\end{tabular}%
}
\end{table}

To understand the specific drivers of end-to-end latency, we profiled the system across a continuous sequence of 100 frames, measuring the temporal cost of each discrete stage in the perception pipeline. Table~\ref{tab:pipeline_latency} provides a side-by-side breakdown of these components, comparing the baseline PyTorch implementation against our compiled TensorRT engine on the Jetson Orin NX.

The system architecture partitions the pipeline into six distinct operational stages: worker handoff, preprocessing, core inference, non-maximum suppression (NMS) and ByteTrack association, outlier filtering, and detection publishing. Under the optimized TensorRT configuration, core inference (7.9,ms) and worker handoff (5.4,ms) constitute the primary latency drivers, together accounting for over 60\% of the 20.8 ms detector-side subtotal. While preprocessing and inference are largely deterministic, the NMS and ByteTrack association stages exhibit non-trivial variance, with max latency reaching 10.8 ms. This variability suggests that multi-object tracking operations occasionally trigger heavy compute re-associations during complex target maneuvers or when multiple candidates require filtering, whereas outlier filtering and publication stages maintain negligible overhead across all test conditions.

\begin{table}[t]
\centering 
\caption{Detection-stream performance during the representative real-world flight.}
\label{tab:real_detection_stream} 
\begin{tabular}{l c} 
\hline 
Metric & Value \\ 
\hline 
Operation duration & 104.53~s \\
Total messages & 3326 \\
Average publish Frame rate & 31.82~FPS \\
Median message gap & 30.29~ms \\
Maximum message gap & 100.72~ms \\
Missing frames & 29 \\
Valid detections & 3065 / 3326 (92.2\%) \\
Mean confidence & 0.813 \\
Large trajectory jumps & 2 \\
\hline 
\end{tabular} 
\end{table}

\begin{table}[t]
\centering
\caption{Detector-side latency comparison: PyTorch vs. TensorRT on Jetson Orin NX.}
\label{tab:pipeline_latency}
\resizebox{\columnwidth}{!}{%
\begin{tabular}{lcccc}
\toprule
 & \multicolumn{2}{c}{\textbf{PyTorch (ms)}} & \multicolumn{2}{c}{\textbf{TensorRT (ms)}} \\
\cmidrule(lr){2-3} \cmidrule(lr){4-5}
\textbf{Stage} & \textbf{Mean} & \textbf{Max} & \textbf{Mean} & \textbf{Max} \\
\midrule
Worker wake / image handoff & 7.2 & 18.1 & 5.4 & 13.7 \\
Preprocessing & 4.1 & 5.8 & 4.8 & 6.3 \\
Inference & 26.1 & 33.6 & 7.9 & 10.6 \\
NMS + ByteTrack association & 7.1 & 12.3 & 2.5 & 10.8 \\
Outlier filtering & $< 0.1$ & 0.1 & $< 0.1$ & $< 0.1$ \\
Detection publishing & $< 0.1$ & 0.1 & 0.2 & 0.3 \\
\midrule
\textbf{Detector-side subtotal} & \textbf{44.5} & \textbf{70.0} & \textbf{20.8} & \textbf{41.7} \\
\bottomrule
\end{tabular}%
}
\vspace{0.5mm}
\begin{flushleft}
\footnotesize{Statistics are computed over the last 100 frames. The subtotal excludes camera SDK and image transport latency. The worst-case bound is the sum of the per-stage maxima.}
\end{flushleft}
\end{table}

The detector-side pipeline (with TensorRT) runs at 48~fps on average on the Jetson Orin NX, excluding camera SDK and image transport overhead. 
This includes image handoff, TensorRT preprocessing and inference, NMS, ByteTrack association, outlier filtering, and detection publishing. These results provide direct evidence that the SCOUT pipeline provides stable visual feedback for closed-loop goalkeeper-style perimeter defense.

\subsection{Discussion}
The realization of an autonomous, air-to-air tracking system operating entirely under edge computational constraints represents a practical alternative to ground-centralized architectures \cite{2017_goppert_cuas}, \cite{2022rudys_hostile_uav}. Recent trends in counter-UAS and search-and-rescue design heavily favor fully onboard inference to bypass ground station telemetry bottlenecks \cite{wyder_2019_dronehunter}, \cite{search_and_rescue}. While vision-based platforms historically rely on high-latency ground telemetry, the SCOUT pipeline demonstrates that integrating optimized TensorRT neural graphs \cite{TensorRT} provides an accelerated, onboard perception solution. Our evaluation indicates that training the YOLO architecture on existing drone detection datasets~\cite{wyder_2019_dronehunter,zheng_2021_detfly,zhao_2022_antiuav,peterson2026_LRDDv3}  effectively yields high visual sensitivity at a distance, but close-quarters proximity introduces severe visual degradation. As the spatial gap closes, rapid scale shifts, abrupt aspect ratio changes, and acute motion blur heavily degrade raw detection confidence, demonstrating that high static dataset accuracy does not inherently translate to localized tracking stability.

The SCOUT architecture mitigates visual dropouts by prioritizing target-center continuity to stabilize the flight controller. By lowering the detector baseline threshold to 0.06 and deploying a dual-pool ByteTrack association strategy \cite{zhang2022bytetrack}, the system recovers low-confidence target boxes against the historical trajectory using a localized Kalman filter. This framework, augmented by sequential geometric filtering, suppresses high-frequency bounding box jitter and false-positive target swaps that frequently plague generic trackers \cite{zheng_2021_detfly}, \cite{zhao_2022_antiuav}. Consequently, the primary systemic benefit of the pipeline is the mitigation of unstable control inputs, transforming fragmented raw detections into a smoothed velocity command stream that preserves qualitative closed-loop goalkeeper behavior during aggressive target maneuvers.

Several critical limitations remain in the current validation framework that must be addressed before deployment in unconstrained environments. First, the experimental evaluation was restricted to a controlled indoor setup using a manually operated intrusion drone, which does not fully replicate the complex trajectory profile of an autonomous adversarial target. Second, while the flight tests demonstrated reliable tracking-stream continuity and responsive physical positioning, the current evaluation lacks aggregate, objective trial-level success metrics like statistical blocking rates or boundary penetration margins across serialized trials. The transition toward self-contained, autonomous UAV architectures enables broader application in fields requiring remote oversight, including search and rescue and infrastructure monitoring \cite{DroneforOil}.

Future work will focus on establishing these quantitative benchmarks in outdoor settings, exploring multi-modal sensor fusion via thermal or acoustic arrays \cite{2019_muller_detection} to handle field-of-view limits, and integrating strict geo-fencing fail-safes to ensure operational safety. Any outdoor testing and deployment for such a counter-UAV system must have a set mechanism to prevent any bystanders from being harmed.

%% file: sections/5.Conclusion.tex
\section{Conclusion}
This work presents a robust, edge-integrated visual perception architecture engineered to overcome the acute challenges of latency, motion blur, and tracking fragmentation in autonomous air-to-air perimeter defense operations. With real-time outlier filtration, algorithmically reclaiming degraded target boxes via a dual-pool ByteTrack association paradigm and a TensorRT accelerated inference engine, the SCOUT pipeline achieves complete operational independence from high-latency ground-station downlink. Experimental validation demonstrates that our standalone edge system delivers a detection rate of 92.2\% while maintaining a highly deterministic inference throughput of 31.85 FPS on embedded hardware. Ultimately, these results prove that localized deep learning frameworks can successfully govern strict real-time control loops required for precise aerial interception, establishing a viable, high-performance blueprint for resource-constrained autonomous defense and navigation platforms.
